\documentclass[conference]{IEEEtran}
\IEEEoverridecommandlockouts
\usepackage{amsmath,amssymb,amsfonts}
\usepackage{algorithm, algorithmic}
\usepackage{textcomp}
\usepackage{mathtools}
\usepackage[table]{xcolor}
\usepackage{xcolor}
\usepackage{microtype}
\usepackage{graphicx}
\usepackage[noadjust]{cite}
\usepackage{booktabs}
\usepackage{hyperref}

\def\BibTeX{{\rm B\kern-.05em{\sc i\kern-.025em b}\kern-.08em
    T\kern-.1667em\lower.7ex\hbox{E}\kern-.125emX}}
\makeatletter
\newcommand{\linebreakand}{%
\end{@IEEEauthorhalign}
\hfill\mbox{}\par
\mbox{}\hfill\begin{@IEEEauthorhalign}
}
\makeatother
\begin{document}

\title{SEVEN: Pruning Transformer Model by Reserving Sentinels\\}

\author{
	\IEEEauthorblockN{1\textsuperscript{st} Jinying Xiao}
	\IEEEauthorblockA{\textit{School of Computer and Communication Engineering} \\
		\textit{Changsha University of Science and Technology}\\
		Changsha, China \\
		xiaojinying1014@163.com}
\\
	\IEEEauthorblockN{3\textsuperscript{rd} Jie Nie}
		\IEEEauthorblockA{\textit{School of Computer and Communication Engineering} \\
		\textit{Changsha University of Science and Technology}\\
		Changsha, China \\
		csustniejie@163.com}
	\and
	\IEEEauthorblockN{2\textsuperscript{nd} Ping Li}
	\IEEEauthorblockA{\textit{School of Computer and Communication Engineering} \\
		\textit{Changsha University of Science and Technology}\\
		Changsha, China \\
		lping9188@163.com}
\\
	\IEEEauthorblockN{4\textsuperscript{th} Zhe Tang}
	\IEEEauthorblockA{\textit{School of Computer and Communication Engineering} \\
		\textit{Changsha University of Science and Technology}\\
		Changsha, China \\
		tangzhe77777@163.com}
}

\maketitle
\begin{abstract}
Large-scale Transformer models (TM) have demonstrated outstanding performance across various tasks. However, their considerable parameter size restricts their applicability, particularly on mobile devices. Due to the dynamic and intricate nature of gradients on TM compared to Convolutional Neural Networks, commonly used pruning methods tend to retain weights with larger gradient noise. This results in pruned models that are sensitive to sparsity and datasets, exhibiting suboptimal performance. Symbolic Descent (SD) is a general approach for training and fine-tuning TM. In this paper, we attempt to describe the noisy batch gradient sequences on TM through the cumulative process of SD. We utilize this design to dynamically assess the importance scores of weights.SEVEN is introduced by us, which particularly favors weights with consistently high sensitivity, i.e., weights with small gradient noise. These weights are tended to be preserved by SEVEN. Extensive experiments on various TM in natural language, question-answering, and image classification domains are conducted to validate the effectiveness of SEVEN. The results demonstrate significant improvements of SEVEN in multiple pruning scenarios and across different sparsity levels. Additionally, SEVEN exhibits robust performance under various fine-tuning strategies. The code is publicly available at https://github.com/xiaojinying/SEVEN.
\end{abstract}


\section{Introduction}
\noindent Pre-trained Transformer models (TM) \cite{ref22}\cite{ref23} offer powerful language representation capabilities and, after fine-tuning, prove effective for various downstream tasks. However, the high performance of these models comes with increased computational costs. Given the substantial growth in the size of TM \cite{ref1}\cite{ref25}, methods to reduce their overhead have become popular. One straightforward approach is pruning \cite{ref26}\cite{ref27}\cite{ref28}, which involves trimming redundant parameters based on a specified criterion. Recent work \cite{ref7} has discovered the existence of "lottery tickets" in TM—subsets of parameters with fewer in number compared to the original network, yet they perform equally well or even better. Fine-grained unstructured pruning is focused on by us, with individual weights in TM being removed to obtain high-performance lottery tickets.

\begin{figure}[tb]
	\centering
	\includegraphics[width=0.48\textwidth, keepaspectratio]{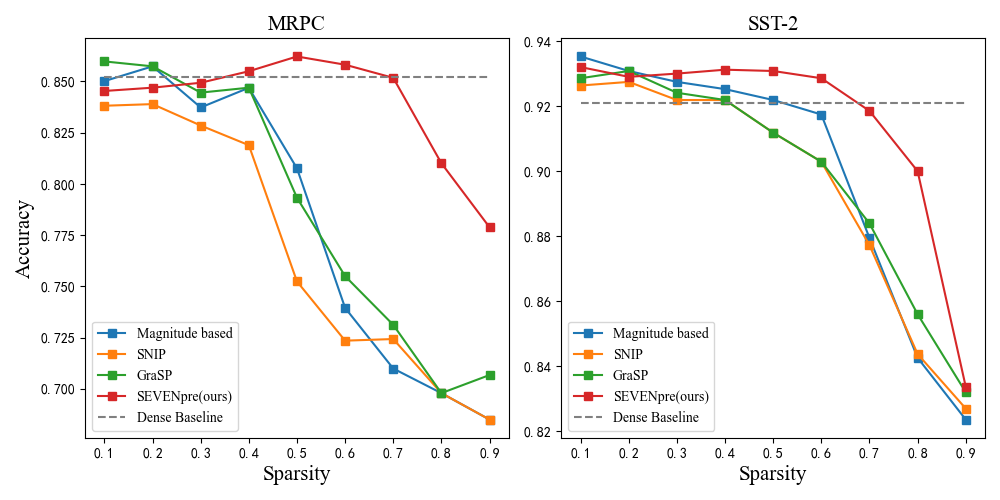}
	\caption{The performance of weight pruning and gradient pruning using BERT-BASE on the MRPC and SST-2 datasets.}
	\label{fig1}
\end{figure}

In unstructured pruning, convolutional neural network (CNN) pruning methods based on gradients or variants as scoring criteria have gained popularity \cite{ref21}\cite{ref8}\cite{ref9}. However, it was surprising to discover that naively applying these methods to TM is impractical, as certain limitations in TM are exhibited by gradient-based methods (SNIP, GrapSP). As illustrated in Fig. \ref{fig1}, although gradient-based pruning methods perform well at high pruning rates, their performance is less satisfactory at moderate sparsity levels. Subsequent experiments also reveal that these gradient-based pruning methods in TM are not universally applicable across multiple datasets, showing effectiveness or ineffectiveness for specific datasets. Thus, it can be inferred that gradient-based methods are not perfect, especially at moderate sparsity levels.In contrast, pre-pruning in TM tends to favor weight pruning methods \cite{ref7}\cite{ref15}. Within the category of gradient-based pruning, TM seem to be more amenable to dynamic pruning \cite{ref9}\cite{ref4}.

\begin{figure}[tb]
	\centering
	\includegraphics[width=0.3\textwidth, keepaspectratio]{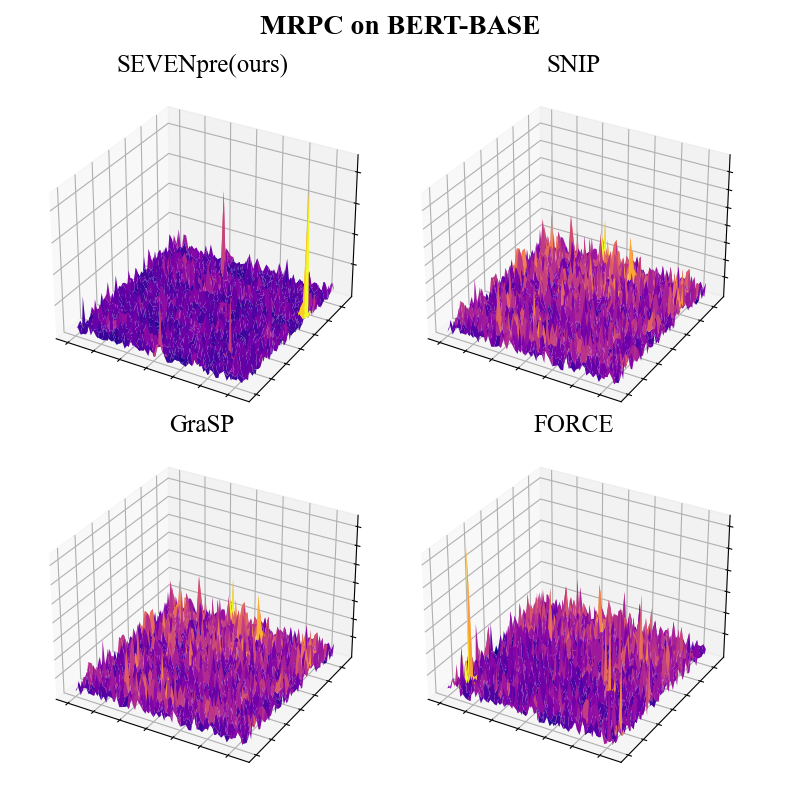}
	\caption{The one-norm of gradient changes for retained weights under a sparsity of 0.5 is calculated. The sampling method involves randomly extracting the norm of the gradient differences for the partially retained weights between consecutive iterations without updates.}
	\label{fig2}
\end{figure}

To address this, the variations in model gradients during the training process are examined. In SGD, the presence of stochastic gradient noise (SGN) is widespread \cite{ref11}. The stochastic descent of gradients implies capturing the learning direction for the current batch only \cite{ref31}, fundamentally influenced by the data itself \cite{ref12}. This phenomenon exists in both CNNs and TM, but in TM, unlike the parameter-sharing convolutional structures in CNNs, gradient behavior is more dynamic and complex \cite{ref29}. Therefore, we believe this dynamic complexity would intensify the impact of noise on pruning. Significant changes in gradient norms for consecutive batches of data are reported after pruning with SNIP, GraSP, and FORCE methods, as shown in Fig. \ref{fig2}. While such situations may be acceptable in single-shot pruning, iterative pruning with the FORCE method also exhibits this problem. This issue becomes more pronounced in text data \cite{ref33}. Thus, on TM, the dynamic and complex gradients may affect the judgment of redundant weights. According to subsequent fine-tuning performance, it is evident that the pruned model's performance is further compromised under the influence of such noise.

Drawing inspiration from the characteristics of Symbolic Descent (SD) \cite{ref36}\cite{ref51}, we propose the SEVEN (re\underline{\textbf{se}}r\underline{\textbf{v}}e s\underline{\textbf{en}}tinels) pruning method, which is heuristic in nature. Specifically, SEVEN considers the impact of SGN during iterative pruning. For weights with high gradients but large noise, termed temporary sentinel weights (TSW), and weights that maintain stable gradients in the long term, termed sentinel weights (SW), SEVEN tends to preserve SW while removing TSW. However, due to the dynamic nature of training, TSW also becomes noise-robust in the later stages \cite{ref33}. Therefore, the introduction of some noise through SD is done to filter out short-term SW. As a result, the model after SEVEN pruning maintains stable gradient changes and robust parameters, as shown in Fig. \ref{fig2}.Furthermore, we dynamically evaluate gradients during iterative pruning, repeating pruning operations to create high-quality lottery subnetworks until the desired sparsity is achieved. It is important to note that SEVEN is a general pruning method for finding high-performance subnetworks. Two versions are proposed: SEVEN\textsubscript{pre}, a pre-pruning method applied before training, and SEVEN\textsubscript{dyn}, a dynamic pruning method performed during training, gradually pruning until the desired sparsity is reached.

Our contributions are as follows:
\begin{itemize}
	\item We investigated the ineffectiveness of transferring efficient gradient pruning methods from CNNs to TM, attributing it to the complexity and dynamism of gradients in TM, specifically the impact of SGN. This situation leads to a preference for retaining TSW with rich SGN during gradient pruning, thereby affecting the subnetwork's performance. These TSW exhibit noticeable gradient changes during subsequent fine-tuning, resulting in suboptimal fine-tuning performance. In contrast, we believe that SW with sustained moderate to high sensitivity and low noise are more worthy of preservation.
	\item We have introduced a general pruning method for finding high-performance subnetworks, named SEVEN, which includes the pre-pruning method SEVEN\textsubscript{pre} and the dynamic pruning method SEVEN\textsubscript{dyn}. SEVEN is a first-order gradient-based pruning approach and does not require rewinding. Specifically, SEVEN dynamically evaluates weights during iterative pruning. For TSW, SEVEN tends to remove them, while for SW, SEVEN leans towards preservation.
    \item We validate SEVEN on various benchmark visual tasks such as CIFAR10/100, ImageNet, and language tasks using the GLUE benchmark, with CLIP and BERT, respectively. Whether it is SEVEN\textsubscript{pre} or SEVEN\textsubscript{dyn}, SEVEN consistently outperforms other state-of-the-art methods, leading by 3\%-6\% on certain datasets. Furthermore, the generality of SEVEN is assessed, examining its pruning timings and fine-tuning strategies. It is found that SEVEN exhibits robust performance across these factors.
\end{itemize}

\section{RELATED WORK}
\noindent Pruning problems are primarily addressed using two major categories of methods \cite{ref40}: unstructured pruning \cite{ref41}\cite{ref42}\cite{ref48}\cite{ref9} and structured pruning \cite{ref43}\cite{ref44}\cite{ref45}\cite{ref49}. Unstructured pruning involves scoring individual weights to remove single parameters from the entire model, while structured pruning involves removing complete modules from the model architecture, such as entire attention heads and layers. Additionally, some mask learning methods directly learn pruning masks during training and use them to prune weights \cite{ref47}.
Theoretical support for model pruning before training is provided by \cite{ref6}\cite{ref7}. Building on this foundation, SNIP \cite{ref21} and GraSP \cite{ref8} explore the potential of single-shot pruning, i.e., pruning only once. Other works \cite{ref5}\cite{ref15}\cite{LNPT} achieve high-performance lottery tickets by performing multiple pruning of the model before training. In TM, pre-pruning is primarily based on weight magnitudes \cite{ref7}\cite{ref15}. This scoring method originates from the inherent properties of pre-trained models and does not carry any fine-tuning-related information for the current task, resulting in suboptimal performance.
In TM, dynamic pruning, carried out concurrently with training, has become mainstream. Movement \cite{ref4}, using first-order derivatives as scoring criteria, achieves good subnetworks at high pruning rates. It proposes a soft version, Soft-Movement, by suppressing the growth of the scoring function. PLATON \cite{ref9} integrates the uncertainty of the scoring criteria to consider redundant weights. However, retaining weights with uncertainty affects the performance of subsequent fine-tuning. In contrast, the aim is to preserve weights that exhibit more stable performance. Before our work, Movement and PLATON methods provided the best performance for unstructured dynamic pruning.

\section{METHOD}
\subsection{Preliminary}
\noindent TM have been widely used for training large neural language models \cite{ref1}\cite{ref2}. It consists of multiple layers with the same structure, each containing a multi-head attention mechanism and one or two feedforward neural networks.The transformer model $f$ parameterized by $\theta$ is represented as $f(\cdot;\theta)$, where $\theta \in \mathbb{R}^{w \times 1}$. The training data is denoted as $\mathcal{D} = (\mathcal{X},\mathcal{Y})$, where $f$ signifies the mapping from the sample space $\mathcal{X}$ to the output space. In the actual training process, the data is divided into $N$ batches, denoted as $(x_i, y_i)$, where $i = 1, 2, \ldots, N$. Consequently, the loss function $\mathcal{L}(f(x_i;\theta))$, abbreviated as $L_i$, is defined for $(x_i, y_i)$.
\subsection{Gradient Noise}
\noindent Efficient pruning using scoring function $\mathcal{S}$ is our objective, however, this task becomes particularly challenging in TM. For a multi-head attention $MultiHead\left(Q,K,V\right)=Concat\left(head_1,\ldots,head_H\right)W^O$, where $head_h=Attention\left(QW_h^Q,KW_h^K,VW_h^V\right)$, and $ Q,K,V\in \mathbb{R}^{n\times d}$, $W_h^Q,W_h^K\in \mathbb{R}^{d\times d_k}$, $W_h^V\in \mathbb{R}^{d\times d_v}$, $W^O\in \mathbb{R}^{d\times h d_v}$.
For the loss on data $\left(x_i,y_i\right)$ denoted as $\mathcal{L}_i$, the gradients with respect to $W_h^Q$ and $W_h^K$ can be expressed as:
\begin{equation}
	\label{eq1}
	\frac{\partial \mathcal{L}_i}{\partial W_h^Q}=\gamma Q^T{\frac{\partial \mathcal{L}_i}{\partial A}}^T\mathbb{P}_h^TKW_H^Q
\end{equation}
\begin{equation}
	\label{eq2}
	\frac{\partial \mathcal{L}_i}{\partial W_h^K}=\gamma K^T\mathbb{P}_h{\frac{\partial \mathcal{L}_i}{\partial A}}^TQW_H^K
\end{equation}

Where, $\gamma=\frac{1}{\sqrt{d_k}}$, $A=\left(\ldots,softmax\left(\frac{QW_H^Q\left(W_H^K\right)^TK^T}{\sqrt{d_k}}\right)\right)$, and $\mathbb{P}_h=\left(\ldots,E_{n\times n}^h,\ldots\right)_{n\times n H}$.

From (\ref{eq1})(\ref{eq2}), it can be observed that for $\frac{\partial \mathcal{L}_i}{\partial W_h^Q}$ (similarly for $\frac{\partial \mathcal{L}_i}{\partial W_h^K}$), the expression of this gradient is not only related to Q and K but also correlated with the gradients of $\mathcal{L}_i$ with respect to other attention heads $\frac{\partial \mathcal{L}_i}{\partial A}$. Thus, in the TM, the expression of gradients is intricately connected globally, making the gradients more dynamic.

To validate our perspective, the Relative Gradient Variations (RGV) during iterations were computed in both CNN and Transformer models. Specifically, for the model's full gradient $\frac{\partial \mathcal{L} \left(f\left(\mathcal{X};\theta\right)\right)}{\partial\theta}$ and the gradient of $\left(x_i,y_i\right)$ denoted as $\frac{\partial \mathcal{L}\left(f\left(x_i;\theta\right)\right)}{\partial\theta}$ (abbreviated as $g$ and $g_i$, where $g,g_i\in \mathbb{R}^{w\times 1})$, we used relative changes in gradients to assess the complexity of the model's gradient variations:
\begin{equation}
	\label{eq3}
	RGV=\frac{g_i-g}{g},RGV\in\mathbb{R}^{w\times1}
\end{equation}

\begin{figure}[tb]
	\centering
	\includegraphics[width=0.48\textwidth, keepaspectratio]{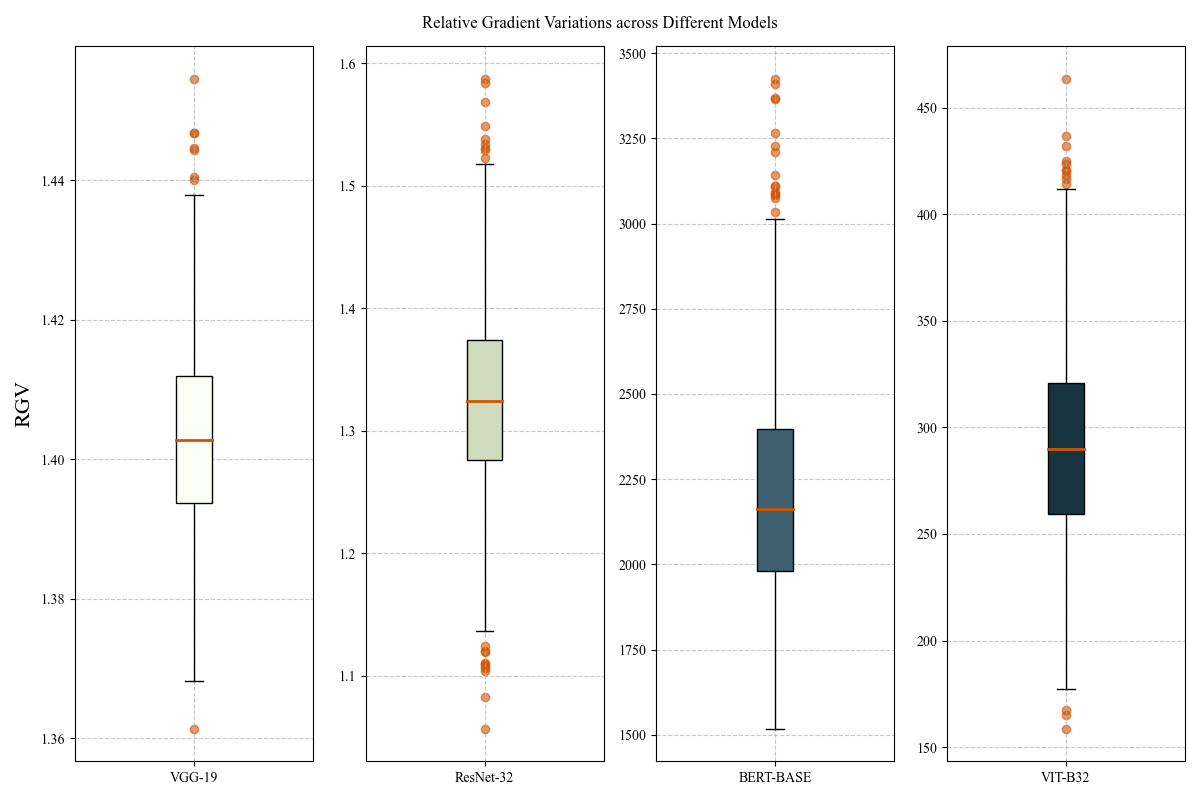}
	\caption{Different distributions of RGV in various networks. BERT-BASE and VIT-B32 represent Transformer models. The sampling method involves iterating using the initial pre-trained model without updating. CIFAR10 dataset was used for VGG, ResNet, and VIT-B32, while SST-2 dataset was used for BERT.}
	\label{fig3}
\end{figure}

Fig. \ref{fig3} reports the 1-norm values of Relative Gradient Variations (RGV) in different network architectures. It can be observed that RGV in TM is significantly larger than in CNNs, and the distribution of RGV in TM is more dispersed with more outliers. This is because TM, unlike the parameter-sharing convolutional structures, make gradient propagation dynamic for each position relative to the input sequence. The output at each position depends on the entire input sequence, making the gradient behavior more complex than in CNNs \cite{ref29}. The belief is that this situation increases the likelihood of TSW occurrences.

Considering the inherent properties of stochastic gradients, due to the randomness of SGD, prior works \cite{ref34}\cite{ref11} have represented SGN as:
\begin{equation}
	\label{eq4}
	N_i=g_i-g,N_i\in\mathbb{R}^{w\times1}
\end{equation}

Most works\cite{ref37}\cite{ref38} assume that $N_i$ follows a Gaussian distribution with covariance matrix $\mathcal{N}\left(0,\sigma\right)$:
\begin{equation}
	\label{eq5}
	\sigma=\frac{1}{S}\left[\frac{1}{N}\sum_{i=1}^{N}{g_i^Tg_i}-g^Tg\right]
\end{equation}

The scoring functions of SNIP and Movement can be expressed as:
\begin{equation}
	\label{eq6}
	\mathcal{S}=\left|\theta\odot N_i+\theta\odot\nabla_\theta \mathcal{L} \left(f\left(\mathcal{X};\theta\right)\right)\right|
\end{equation}

It can be observed that due to the noise, the direction and magnitude of stochastic gradient pruning and the full gradient are inconsistent, with strong batch sampling randomness \cite{ref36}, and \cite{ref33}\cite{ref34}\cite{ref39} indicates that the SGN has heavy tails, far exceeding their averages.
Therefore, for the scoring function $\mathcal{S}$, noise $N_i$ is a crucial issue affecting pruning performance. Combining (\ref{eq3}) and (\ref{eq5}), we get:
\begin{equation}
	\label{eq7}
	\begin{split}
		\sigma=&\frac{1}{S}[\frac{1}{N}\sum_{i=1}^{N}{\left(RGV\odot g\right)^T\left(RGV\odot g\right)}\\
		&+\left(RGV\odot g\right)^Tg+g^T\left(RGV\odot g\right)]
	\end{split}
\end{equation}

From the (\ref{eq7}), it can be observed that for larger RGV values, we obtain gradients with larger variances, indicating significant SGN.

While SGN aids the model in escaping local optima, benefiting SGD \cite{ref34}\cite{ref35}, recent works \cite{ref51}\cite{ref36} suggest that in TM, adaptive optimization algorithms with low noise perform better. The inference is that pruning methods based on $\mathcal{S}$ in TM may not be accurate due to the significant impact of SGN. $\mathcal{S}$ is likely to favor retaining TSW, thereby influencing the subsequent fine-tuning of the model, as shown in Fig. \ref{fig1}.

\section{SEVEN}
\noindent Next, we introduce the pruning method SEVEN, inspired by gradient clipping \cite{ref52}\cite{ref33}. Gradient clipping suppresses the impact of noise by scaling the gradients to a certain range. The aim is to apply a similar correction, akin to clipping, to $\nabla_{\theta}\mathcal{L} \left(f\left(x_i;\theta\right)\right)$ in the environment of SGN.

\cite{ref33} indicates that adaptive optimization algorithms perform better in TM than SGD, primarily due to the presence of Stochastic SGN. However, from previous experimental results \cite{ref36}, the performance of SGD in a denoising environment did not show a significant improvement. Therefore, it can be explicitly considered that using the full gradient for pruning is suboptimal. In a denoising environment, SD and its variants outperform Adam \cite{ref36}\cite{ref51}, partly because SD introduces imprecise gradients during weight updates. In contrast, in an SGN environment, Adam performs better. Due to the training dynamics of weights, we cannot assert that SW maintain robustness throughout the entire training cycle. There is every reason to believe that a certain level of SGN is necessary, and, as a result, the use of stochastic gradients continues. From the results, Polyak averaging plays a significant role in SD \cite{ref36}\cite{ref51}\cite{ref50}. Consequently, SEVEN combines SD with Polak averaging to modify and evaluate the stochastic gradient in a manner similar to Adam.

For ease of notation, let's denote $\nabla_{\theta} \mathcal{L} \left(f\left(x_i;\theta\right)\right)$ as $g_i$ where $g_i\in \mathbb{R}^{w\times 1}$. Based on empirical observations, we process $g_i$ and $g_i^2$ using Polyak averaging:

\begin{equation}
	\label{eq8}
	\overline{g_i}=\alpha_1\overline{g_{i-1}}+\left(1-\alpha_1\right)g_i
\end{equation}
\begin{equation}
	\label{eq9}
	\overline{g_i^2}=\alpha_2\overline{g_{i-1}^2}+\left(1-\alpha_2\right)g_i^2
\end{equation}

Where $\alpha_1$,$\alpha_2$ are hyperparameters in the range $\left(0,1\right)$, and $\overline{g_0}=\overline{g_0^2}=0$.
A small subset of gradients from the original data is selected, denoted as $S=\{g_1,g_2,\ldots,g_n\}$, where $n<N$. Additionally, unbiased correction is applied to $\bar{g_i}$ and $\bar{g_i^2}$:
\begin{equation}
	\label{eq10}
	\hat{\mu}_i^{\left(1\right)}=\frac{\overline{g_i}}{1-\alpha_1^i}
\end{equation}
\begin{equation}
	\label{eq11}
	\hat{\mu}_i^{\left(2\right)}=\sqrt{\frac{\overline{g_i^2}}{1-\alpha_2^i}+\epsilon}
\end{equation}

Where $\epsilon$ is a very small number to prevent division by zero.

According to Symbolic Descent (SD), the correction coefficients for $g_i$ are given by:
\begin{equation}
	\label{eq12}
	\hat{\mu}=\hat{\mu}_i^{\left(1\right)}/\hat{\mu}_i^{\left(2\right)}
\end{equation}

We perform element-wise multiplication (Hadamard product) between $\hat{\mu}$ and the current gradient $g_i$. From the formula, $\hat{\mu}$ provides correction to $g_i$:
\begin{equation}
	\label{eq13}
	\hat{g}_i=g_i\odot\hat{\mu}
\end{equation}

The metric for $\hat{g}_i$ not only reflects our goal of preserving weights with the maximum gradient but also suppresses weights sensitive to noise by measuring the gradient magnitude through the second moment. The scoring function is constructed in the form of a synapse \cite{ref10}:
\begin{equation}
	\label{eq14}
	\mathcal{S}_i=\left|\theta\odot\hat{g}_i\right|
\end{equation}

We accumulate $\mathcal{S}_i$, and the scoring function is represented as:
\begin{equation}
	\label{eq15}
	\mathcal{S}=\sum_{i=1}^{n}\left|\theta \odot\hat{g}_i\right|
\end{equation}
\begin{algorithm}[tb]
	\caption{SEVEN\textsubscript{pre}}
	\label{alorithm pre}
	\begin{algorithmic}[1]
		\REQUIRE pre-trained model: $\{\theta_0\}$; train data: $\mathcal{D}$; sparsity: $s$; total number of iterations $T$; pruning steps: $K$.
		\hrule
		\vspace{0.1cm}
		\FOR{$t$ from 1 to $T$}
		\IF{$t<=K$}
			\STATE Update $\bar{g_t}$, $\bar{g_t^2}$(see eq 8,9)
			\STATE Computer the score $\mathcal{S}\left(\theta_t\right)$(see eq 15)
			\STATE Computer the pruning rate for the current step $ 1-(1-s)^\frac{t}{K}$ as $\mathcal{P}$
			\STATE Compute $\mathcal{P}_{th}$ percentile of $\mathcal{S}\left(\theta_t\right)$ as $\tau$
			\STATE $\mathcal{M}=\mathcal{S}\left(\theta_t\right)<\tau$
			\STATE Mask model: $\{\theta_{t}\odot{\mathcal{M}\}}$
		\ENDIF
		\STATE Update model: $\theta_{t+1}=\theta_t-\eta\nabla_{\theta_t} \mathcal{L} \left(\theta_t\right)$
		
		\ENDFOR
		\vspace{0.1cm}
		\hrule
		\vspace{0.1cm}
		\RETURN  model:$\{\theta\odot{\mathcal{M}\}}$
	\end{algorithmic}
\end{algorithm}

\begin{algorithm}[tb]
	\caption{SEVEN\textsubscript{dyn}}
	\label{alorithm dyn}
	\begin{algorithmic}[1]
		\REQUIRE pre-trained model: $\{\theta_0\}$; train data: $\mathcal{D}$; sparsity: $s$; pruning steps: $K$, total number of iterations $T$; iteration of pruning start $t_i$.
		\vspace{0.1cm}
		\hrule
		\vspace{0.1cm}
		\FOR{$t$ from 1 to $T$}
		
		\IF{$t>t_i$ and $t\le t_i+K$}
			\STATE Update $\bar{g_t}$, $\bar{g_t^2}$(see eq 8,9)
			\STATE Computer the score $\mathcal{S}\left(\theta_t\right)$(see eq 15)
			\STATE Computer the pruning rate for the current step $s-s\times\left(1-\left(\frac{t-t_i}{K}\right)^3\right)$ as $\mathcal{P}$
			\STATE Compute $\mathcal{P}_{th}$ percentile of $\mathcal{S}\left(\theta_t\right)$ as $\tau$
			\STATE $\mathcal{M}=\mathcal{S}\left(\theta_t\right)<\tau$
			\STATE Mask model: $\{\theta_{t+1}\odot{\mathcal{M}\}}$
		\ENDIF
		\STATE Update model: $\theta_{t+1}=\theta_t-\eta\nabla_{\theta_t}\mathcal{L} \left(\theta_t\right)$

		\ENDFOR
		\vspace{0.1cm}
		\hrule
		\vspace{0.1cm}
		\RETURN  model:$\{\theta\odot{\mathcal{M}\}}$
	\end{algorithmic}
\end{algorithm}
It can be observed that not only does (\ref{eq14}) take into account the metric for stochastic gradients, but (\ref{eq15}) also accumulates stochastic gradients and their correction coefficients, preserving the stochastic gradients during the SD process. This is different from the training process with SD. Consequently, $\mathcal{S}$ can objectively reflect the actual sensitivity of the current weights, enabling the identification of SW and TSW.
The specific method for SEVEN\textsubscript{pre} is detailed in Algorithm \ref{alorithm pre}.

Next, let's discuss the dynamic pruning method, SEVEN\textsubscript{dyn}. In dynamic pruning, pruning is typically performed after a certain iteration $t_i$, and as fine-tuning progresses, pruning is carried out $K$ times until the desired sparsity level is reached, followed by next fine-tuning. It is worth noting that while SEVEN\textsubscript{pre} utilizes an exponential schedule for the sparsity ratio, SEVEN\textsubscript{dyn}, inspired by \cite{ref9}, employs a cubic schedule. The detailed method for SEVEN\textsubscript{dyn} is provided in Algorithm \ref{alorithm dyn}.

\section{EXPERIMENTS}
\noindent In our experiments, the pre-trained models provided by the official CLIP \cite{ref14} and the pre-trained BERT-BASE model provided by HuggingFace \cite{ref16} were utilized. For fine-tuning CLIP, the classification layer was frozen to avoid introducing any unnecessary learnable parameters. Various image classification tasks from \cite{ref17}, including GTSRB, MNIST, SVHN, CIFAR 10/100, and the large dataset ImageNet, were considered. In the BERT experiments, a task-specific classification layer was added for each task, constituting approximately 3\% of the total parameters. The GLUE benchmark \cite{ref18} was used for evaluation, following the standards outlined in \cite{ref7}. Additionally, the SQuAD question-answering task was examined. It's worth noting that, due to the embedding layer involving only query operations, the embedding layer was ignored during pruning, and the focus was on the critical training components. Our method was compared with recent pruning techniques such as Lottery Ticket \cite{ref19}, ISP \cite{ref15}, SNIP \cite{ref21}, GraSP \cite{ref8}, FORCE \cite{ref5}, PLATON \cite{ref9}, and others.

\subsection{Pre-pruning Method Comparison}
\noindent In this section, pre-trained models for CLIP (VIT-B32) \cite{ref17} and BERT (base) \cite{ref22} were fine-tuned before training. CLIP is supervised by contrasting images and text, determining whether an image and text pair belong together. For effective comparison, all baselines and pruning methods were trained according to the hyperparameters in Table \ref{table1} and implemented following the provided hyperparameters in the respective papers. The hyperparameters for SEVEN\textsubscript{pre} are reported in Appendix \ref{appB}.

The results of SEVEN\textsubscript{pre} on CLIP are presented in Table \ref{table1}. Firstly, it can be observed that the expensive LTH generally outperforms single-shot methods like SNIP and GraSP, and the rewind of the Lottery Ticket Hypothesis indeed brings some improvement, though not significantly. SEVEN is highly effective. Specifically, on the CIFAR10 dataset at an 80\% sparsity, it achieved an improvement of 3.33\%, and similarly, on the CIFAR100 dataset at an 80\% sparsity, it achieved an improvement of 3.21\%. Moreover, on the SVHN dataset, the improvement is even more significant at high sparsity, reaching 6.42\%. Overall, our method performs better as the compression rate increases.

Table \ref{table2} reports the performance in BERT, and unsurprisingly, our method outperforms others in various datasets with a significant performance improvement. It's noteworthy that our method surpasses dense models in these tasks, not only because the model size is reduced, reducing overfitting, but also because our method removes TSW, resulting in better performance for retained weights.

\begin{figure}[tb]
	\centering
	\includegraphics[width=0.48\textwidth, keepaspectratio]{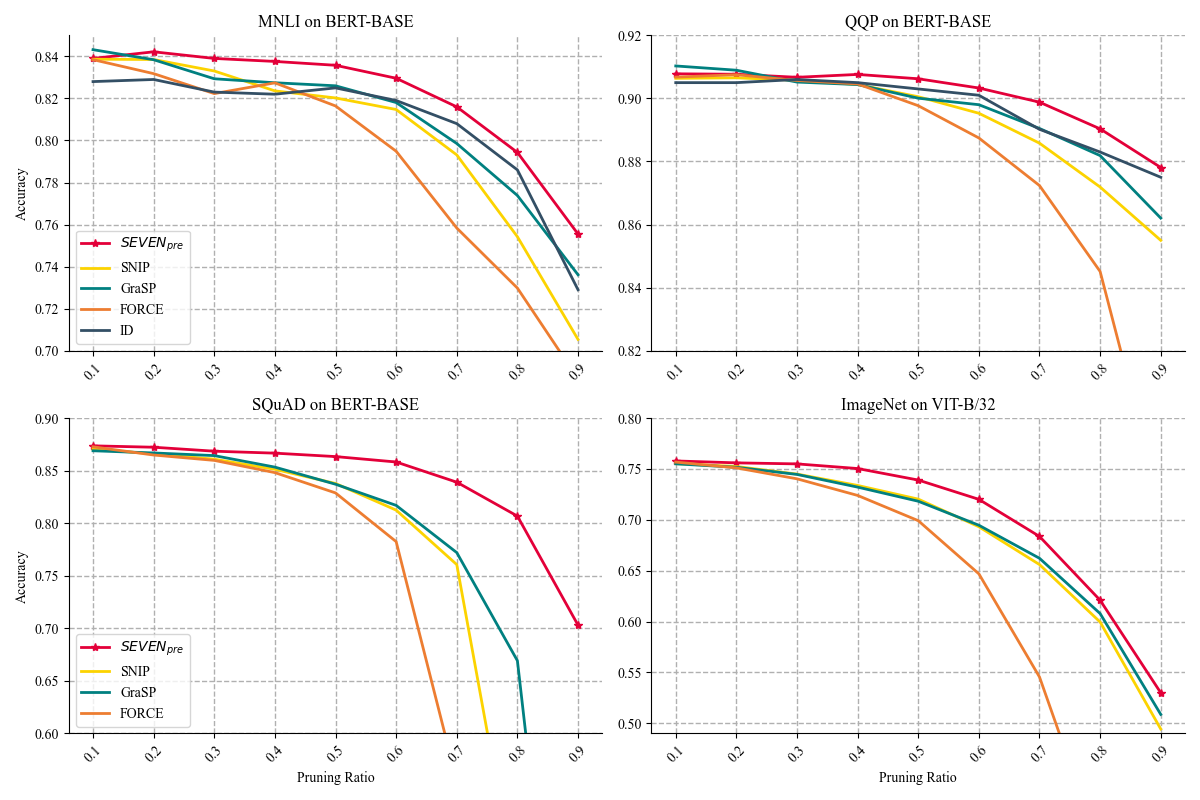}
	\caption{Comparison of different pre-pruning methods on large data sets}
	\label{fig4}
\end{figure}

To further explore the superior performance of SEVEN\textsubscript{pre}, experiments were conducted on large datasets. For CLIP, the ImageNet dataset with over 10 million images was utilized, and for the BERT model, the larger-scale MNLI, QQP, and SQuAD datasets for question-answering were used. The results are reported in Fig. \ref{fig4}, demonstrating that SEVEN\textsubscript{pre} continues to lead on large datasets. It's worth noting that ID (Information Deletion) \cite{ref30} is a pre-pruning method applicable to BERT.
\begin{table}[tb]
	\centering
	\caption{Performance of Different Pruning Methods on Various Datasets in CLIP (VIT-B32)}
	\label{table1}
	\resizebox{\columnwidth}{!}{
		\begin{tabular}{lcccccccccccccccc}
			\toprule
			Dataset & \multicolumn{3}{c}{MNIST} & \multicolumn{3}{c}{SVHN} & \multicolumn{3}{c}{GTSRB} & \multicolumn{3}{c}{CIFAR10} & \multicolumn{3}{c}{CIFAR100} \\
			\cmidrule(lr){1-1} \cmidrule(lr){2-4} \cmidrule(lr){5-7} \cmidrule(lr){8-10} \cmidrule(lr){11-13} \cmidrule(lr){14-16}
			Pruning ratio & 70\% & 80\% & 90\% & 70\% & 80\% & 90\% & 50\% & 60\% & 70\% & 60\% & 70\% & 80\% & 60\% & 70\% & 80\% \\
			\midrule
			Full CLIPViT-B32 & &99.61 & & &97.40 & & &99.08 & & &97.60 & & & 89.35 \\
			\midrule
			Random & 98.71 & 97.42 & 87.04 & 89.85 & 85.61 & 73.74 & 93.76 & 93.65 & 90.97 & 74.51 & 69.84 & 64.96 & 45.20 & 39.92 & 43.24 \\
			One-shot[Mag] & 99.27 & 98.87 & 97.47 & 95.02 & 91.96 & 85.76 & 98.60 & 97.84 & 96.14 & 95.31 & 86.56 & 75.85 & 80.39 & 63.18 & 46.63 \\
			Progressive[Mag] & 99.52 & 97.77 & 95.19 & 95.78 & 90.53 & 85.75 & 98.97 & 97.57 & 96.34 & 95.25 & 90.87 & 78.11 & 81.79 & 70.53 & 60.93 \\
			EarlyBird & 99.38 & 98.96 & 97.64 & 96.34 & 95.93 & 87.02 & 98.15 & 98.19 & 97.26 & 96.06 & 94.18 & 86.84 & 84.22 & 76.79 & 65.67 \\
			SNIP & 99.25 & 98.72 & 97.50 & 95.33 & 91.94 & 82.98 & 98.62 & 97.95 & 96.22 & 95.01 & 87.45 & 76.12 & 81.10 & 62.89 & 55.89 \\
			GraSP & 99.30 & 98.51 & 97.15 & 95.09 & 91.44 & 84.72 & 98.37 & 97.42 & 95.91 & 95.20 & 86.89 & 75.88 & 80.67 & 66.31 & 52.30 \\
			LTH & 99.41 & 99.38 & 98.22 & 96.69 & 95.28 & 87.41 & 98.71 & 98.35 & 97.79 & 96.42 & 94.91 & 87.47 & 84.25 & 78.60 & 65.38 \\
			LTH-Rewind & 99.62 & \textbf{99.64} & 98.18 & 96.72 & 95.22 & 87.47 & \textbf{98.78} & 98.36 & 97.87 & 96.53 & 94.88 & 87.28 & 84.46 & 78.62 & 65.71 \\
			Lottery Pool & 99.25 & 98.97 & 97.76 & 96.54 & 95.12 & 87.29 & 98.52 & 98.30 & 97.55 & 96.14 & 94.50 & 87.11 & 84.07 & 78.21 & 64.39 \\
			SEVEN\textsubscript{pre}(ours) & \textbf{99.65} & 99.60 & \textbf{99.27} & \textbf{97.03} & \textbf{96.36} & \textbf{93.89} & 98.57 & \textbf{98.59} & \textbf{98.17} & \textbf{97.61} & \textbf{96.41} & \textbf{90.61} & \textbf{85.70} & \textbf{81.15} & \textbf{68.92} \\
			\bottomrule
	\end{tabular}}
\end{table}

\begin{table}[tb]
	\centering
	\caption{Performance of Different PRE-Pruning Methods on GLUE Tasks in BERT-BASE}
	\label{table2}
	\resizebox{\columnwidth}{!}{
		\begin{tabular}{lccccccc}
			\toprule
			Dataset & STS-B & WNLI & QNLI & MRPC & RTE & SST-2 & CoLA \\
			\midrule
			Pruning ratio & 50\% & 90\% & 70\% & 50\% & 60\% & 60\% & 50\% \\
			\midrule
			Full BERTBASE & 88.4 & 60.2 & 89.1 & 85.2 & 69.3 & 92.1 & 58.3 \\
			Random & 21 & 53.5 & 61.9 & 69.6 & 56 & 83.1 & 9.6 \\
			One-shot & 83.9 & 53.1 & 86.2 & 83.7 & 62.9 & 86.5 & 49.7 \\
			Progressive & 85 & 53.3 & 87.2 & 83.8 & 65.4 & 86.6 & 52.2 \\
			EarlyBird & 88.1 & 54 & 88.5 & 84.6 & 66.1 & 91.2 & 53.5 \\
			Lottery Ticket & 88.2 & 54.9 & 88.9 & 84.9 & 65 & 91.9 & 53.8 \\
			Lottery Pool & 86.4 & 50.9 & 87.6 & 84.5 & 62.7 & 90.9 & 52.6 \\
			ISP&88.6&55.3&90&85.4&66&92.4&53.6\\
			SEVEN\textsubscript{pre}(ours) & \textbf{89.7} & \textbf{60.3} & \textbf{90.5} & \textbf{86.2} & \textbf{70.5} & \textbf{92.9} & \textbf{58.2} \\
			\bottomrule
		\end{tabular}
	}
\end{table}

\begin{table}[t]
	\centering
	\caption{Comparison of Dynamic Pruning Methods on GLUE}
	\label{table3}
	\resizebox{\columnwidth}{!}{
		\begin{tabular}{lcccccccc}
			\toprule
			Dataset & MNLI & RTE & QNLI & MRPC & QQP & SST-2 & CoLA & STS-B \\
			\midrule
			Pruning ratio & 70\% & 60\% & 70\% & 50\% & 90\% & 60\% & 50\% & 50\% \\
			\midrule
			Full BERT-BASE & 84.3 & 69.3 & 89.1 & 85.2 & 91.1 & 92.1 & 58.3 & 88.4 \\
			Movement & 81.14 & 71.18 & 87.94 & 86.27 & 88.83 & 91.40 & 56.02 & 89.65 \\
			Soft-Movement & 81.89 & \textbf{72.92} & 88.56 & 85.54 & 89.11 & 91.56 & 55.93 & 89.57 \\
			PLATON & 83.32 & 71.39 & 90.54 & 86.52 & 90.27 & 92.41 & 57.18 & \textbf{90.19} \\
			SEVEN\textsubscript{dyn}(ours) & \textbf{83.34} & 72.15 & \textbf{91.04} & \textbf{87.74} & \textbf{90.37} & \textbf{92.75} & \textbf{58.21} & 90.17 \\
			\bottomrule
		\end{tabular}
	}
\end{table}

\begin{table}[tb]
	\centering
	\caption{Impact of Score Function Variants on Pruning Results}
	\label{table4}
	\resizebox{\columnwidth}{!}{
		\begin{tabular}{|c|c|c|c|c|c|c|}
			\hline
			Pruning ratio & \multicolumn{3}{c|}{70\%} & \multicolumn{3}{c|}{60\%} \\ \hline
			Dataset & CIFAR10 & CIFAR100 & QNLI & CIFAR10 & CIFAR10* & SST-2 \\ \hline
			$\hat{\mu}=\hat{\mu}_i^{\left(1\right)}$ & 93.53 & 73.30 & 87.3 & 96.28 & 80.76 & 91.2 \\ \hline
			$\hat{\mu}=\hat{\mu}_i^{\left(2\right)}$ & 93.80 & 73.77 & 86.7 & 96.29 & 80.64 & 91.1 \\ \hline
			$\hat{\mu}=\hat{\mu}_i^{\left(1\right)}\odot\hat{\mu}_i^{\left(2\right)}$  & 94.67 & 76.63 & 86.5 & 95.60 & 77.44 & 90.1 \\ \hline
			\hiderowcolors
			\rowcolor{gray!50}
			SEVEN\textsubscript{pre} & 96.41 & 81.15 & 89.8 & 97.61 & 85.70 & 92.9 \\ \hline
	\end{tabular}}
\end{table}
\subsection{Dynamic Pruning Method Pruning}
To demonstrate the high performance of SEVEN\textsubscript{dyn}, a comparison was made with dynamic pruning methods on BERT, including PLATON \cite{ref9}, Movement \cite{ref4}, and Soft-Movement. The provided hyperparameters for pruning and fine-tuning in each paper were strictly followed. It's important to note that the hyperparameter settings for SEVEN\textsubscript{dyn} and SEVEN\textsubscript{pre} are different, and the hyperparameters for SEVEN\textsubscript{dyn} are reported in Appendix \ref{appB}.

Table \ref{table3} reports the results of obtaining high-performance subnetworks with different dynamic pruning methods. From the results, SEVEN\textsubscript{dyn} not only outperforms the original network but is also more effective than other pruning methods. Specifically, SEVEN\textsubscript{dyn} achieves nearly a 2\% improvement over the full network on the STS-B dataset and a 1.22\% improvement over PLATON on the MRPC dataset. Therefore, we consider SEVEN\textsubscript{dyn} to be highly effective.

\subsection{Analysis}
\begin{figure}[tb]
	\centering
	\includegraphics[width=0.48\textwidth, keepaspectratio]{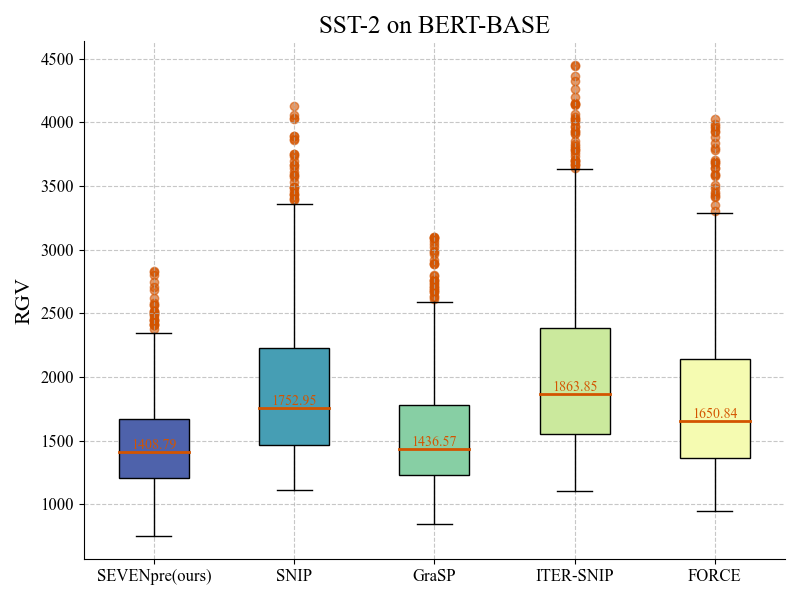}
	\caption{Using the BERT-BASE model with 60\% sparsity, we examined the distribution of the L1 norm of RGV (Relative Gradient Value) across different pre-pruning methods on the SST-2 dataset. The sampling method involves iterating using the initial pre-trained model without updating.}
	\label{fig5}
\end{figure}

\noindent \textbf{SW or TSW?} We advocate maintaining the stability of the model's gradient changes by preserving more SW to achieve better generalization performance. Fig. \ref{fig5} reports the distribution of RGV across multiple batches for different pre-pruning pruning methods. Various gradient pruning methods were examined, and it's evident that in the SEVEN method, the overall values of RGV are lower, and the distribution is relatively concentrated. It can be concluded that our method indeed preserves as many SW as possible.

\begin{figure}[tb]
	\centering
	\includegraphics[width=0.48\textwidth, keepaspectratio]{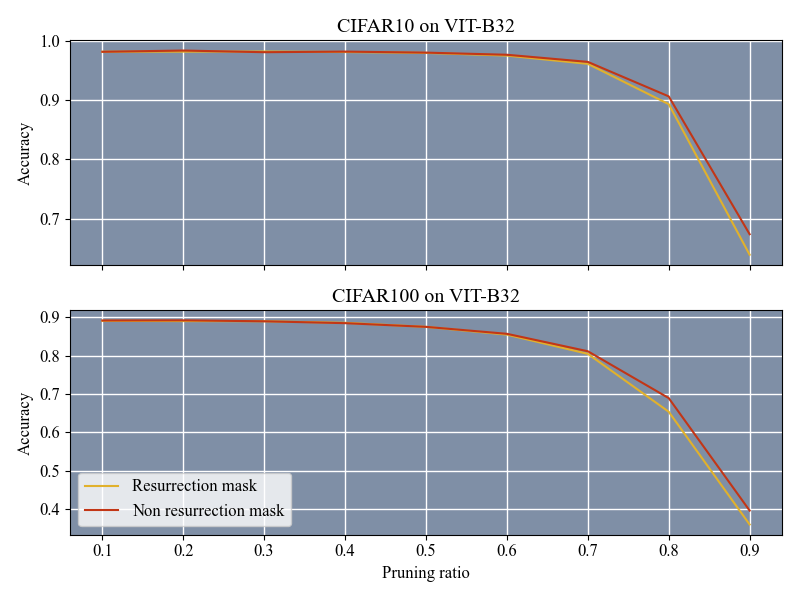}
	\caption{The Impact of Mask Resurrection on SEVEN\textsubscript{pre} Performance}
	\label{fig6}
\end{figure}

\textbf{Mask resurrection.} In \cite{ref5}, the issue of whether the mask is resurrected during iterative pruning is primarily discussed. The article suggests resurrecting weights that have been pruned. In our view, this is a comprehensive approach to consider weights. Consequently, we further investigate the impact of mask resurrection on our method. Fig. \ref{fig6} illustrates the influence of mask resurrection on SEVEN\textsubscript{pre}. It can be observed that our method is robust at low sparsity levels, and mask resurrection has minimal impact on SEVEN\textsubscript{pre}. However, at high sparsity levels, not resurrecting the mask results in better accuracy, indicating that SEVEN\textsubscript{pre} is not influenced by SGN, and the pruned weights are indeed redundant. No further resurrection or adjustment is needed, making SEVEN\textsubscript{pre} reassuringly free from mask resurrection requirements.

\textbf{Score function optimization.} Similar to \cite{ref9}, for experimental validation of our method, an ablation study was conducted. Specifically, our method evaluates the ratio of historical gradients to the gradient spread. We explored various variants of (\ref{eq12}), such as using only $\hat{\mu}_i^{(1)}$, $\hat{\mu}_i^{(2)}$, and $\hat{\mu}_i^{(1)}\odot \hat{\mu}_i^{(2)}$. Table \ref{table4} reports the performance of these variants of the score function. It is evident that our method still leads. In particular, at 70\% sparsity, our method outperforms by 2\%-8\%, and at 60\% sparsity, it leads by 1\%-5\%.
\begin{table}[tb]
	\centering
	\caption{performance of the SEVEN\textsubscript{pre} pruning method using Lion fine-tuning on different datasets}
	\label{table5}
	\begin{tabular}{lccccc}
		\hline
		Dataset & SST-2 & MRPC & CoLA & STS-B & RTE \\
		\hline
		Sparsity & 60\% & 50\% & 50\% & 50\% & 60\% \\
		\hline
		Full BERT-BASE & 92.1 & 85.2 & 58.3 & 88.4 & 69.3 \\
		\hline
		Lottery Ticket & 91.9 & 84.9 & 53.8 & \textbf{88.2} & 65 \\
		EarlyBird & 91.2 & 84.6 & 53.5 & 88.1 & 66.1 \\
		SEVEN\textsubscript{pre}-Lion & \textbf{92.2} & \textbf{85.6} & \textbf{55.5} & 88.1 & \textbf{66.3} \\
		\hline
	\end{tabular}
\end{table}

\textbf{Generalization to Fine-Tuning.} To further validate the generality of our pruning criterion, we conducted fine-tuning using the optimizer Lion \cite{ref53}, which is based on automated search. The results are reported in Table \ref{table5}. It can be observed that with a fine-tuning method derived from thousands of TPU hours of computational search and human intervention, the performance of the pruned model remains superior. Our pruning criterion proves to be robust across various fine-tuning methods, indicating the inherent high performance of the pruned model.

\section{Conclusion}
We propose the TM pruning algorithm SEVEN, which includes the pre-pruning method SEVEN\textsubscript{pre} and the dynamic pruning method SEVEN\textsubscript{dyn}. It dynamically considers the distribution of SGN in the model. In SEVEN, inspired by gradient clipping, we correct the stochastic gradient. In addition, we describe the noisy batch gradient sequence as an accumulation process with sign descent. From the results, these methods retain the stably graded SW and achieve good performance. We conducted extensive experiments on natural language understanding, question answering, and image classification. The results show that SEVEN is significantly superior to existing methods. Our approach can obtain well-performing subnetworks, such as in BERT-BASE, where performance on the QNLI dataset is 2\% higher at 70\% sparsity compared to dense networks. Furthermore, experimental verification shows that SEVEN does not require the resurrection of pruned weights and is robust to various fine-tuning methods.

\bibliographystyle{IEEEtran}

\bibliography{bibtex}

~\\
\appendices

\setcounter{table}{0}   
\setcounter{figure}{0}
\setcounter{section}{0}
\setcounter{equation}{0}
\vspace{-\baselineskip}
\section{}\label{appB}
\renewcommand{\thetable}{B\arabic{table}}
\renewcommand{\thefigure}{B\arabic{figure}}
\renewcommand{\thesection}{B\arabic{section}}
\renewcommand{\theequation}{B\arabic{equation}}

\subsection{SEVEN\textsubscript{pre} Hyperparameter Details}\label{appBA}

For GLUE tasks, following the configuration in \cite{ref7}, we set the epoch to 3, batch size to 32, learning rate to 2e-5, and optimizer to Adam for each dataset. Table \ref{table6} lists the other hyperparameter settings for each dataset in the GLUE tasks. For different pruning rates, $\alpha_1$ and $\alpha_2$ remain unchanged.

\begin{table}[h]
	\centering
	\caption{BERT-BASE Pre-pruning Hyperparameters on GLUE Tasks}
	\label{table6}
	\begin{tabular}{lccccc}
		\hline
		& MNLI & RTE & QNLI & MRPC & QQP \\
		\hline
		$K$ & 150 & 77 & 100 & 100 & 150  \\
		$\alpha_1$ & 0.1 & 0.8 & 0.1 & 0.8 & 0.1  \\
		$\alpha_2$ & 0.2 & 0.9 & 0.2 & 0.9 & 0.1  \\
		\hline
		& SST-2 & CoLA & STS-B & WNLI\\
		\hline
		$K$ & 100 & 100 & 100 & 20 \\
		$\alpha_1$ & 0.5 & 0.8 & 0.8 & 0.8 \\
		$\alpha_2$ & 0.7 & 0.9 & 0.9 & 0.9 \\
		\hline
	\end{tabular}
\end{table}

On the SQuAD dataset, fine-tuning was conducted for only 3 epochs, with a learning rate set to 3e-5, batch size set to 16, and $\alpha_1$, $\alpha_2$, and $K$ set to 0.8, 0.9, and 100, respectively. 

For CLIP, the VIT-B32 model was fine-tuned for 10 epochs with a learning rate of 2e-5. The Adam optimizer was used, 500 iterations for warm-up, and cosine decay was employed. In the pruning settings, $K$ was set to 100, and other hyperparameters are detailed in Table \ref{table7}.

\begin{table}[h]
	\centering
	\caption{VIT-B32 Pre-pruning Hyperparameters on Various Datasets}
	\label{table7}
	\begin{tabular}{lcccccc}
		\hline
		 & MNIST & SVHN & GTSRB & CIFAR10 & CIFAR100 & ImageNet \\
		\hline
		$\alpha_1$ & 0.8 & 0.8 & 0.8 & 0.8 & 0.8 & 0.8 \\
		$\alpha_2$ & 0.9 & 0.9 & 0.9 & 0.9 & 0.9 & 0.9 \\
		\hline
	\end{tabular}
\end{table}
\subsection{SEVEN\textsubscript{dyn} Hyperparameter Details}\label{appBB}
For GLUE tasks, SEVEN\textsubscript{dyn} has a learning rate of 2e-5, a batch size of 32, utilizes the Adam optimizer, and is trained for 10 epochs. These parameters are consistent with the methods we are comparing against. It is worth noting that, in addition to $\alpha_1$ and $\alpha_2$, the pruning iteration count $K$ and the pruning start iteration count $t_i$ are also included. Table \ref{table8} reports these hyperparameters.

\begin{table}[h]
	\centering
	\caption{BERT-BASE Dynamic Pruning Hyperparameters on GLUE Tasks}
	\label{table8}
	
	\begin{tabular}{lcccc}
		\hline
		& MNLI & RTE & QNLI & MRPC  \\
		\hline
		$t_i$ & 5400 & 200 & 2000 & 300  \\
		$K$ & 16600 & 1000 & 10000 & 600  \\
		$\alpha_1$ & 0.7 & 0.8 & 0.8 & 0.7  \\
		$\alpha_2$ & 0.9 & 0.8 & 0.9 & 0.9  \\
		\hline
		& QQP & SST-2 & CoLA & STS-B \\
		\hline
		$t_i$ & 5400 & 1000 & 500 & 500 \\
		$K$ & 16600 & 4000 & 1000 & 2000 \\
		$\alpha_1$ & 0.7 & 0.8 & 0.8 & 0.8 \\
		$\alpha_2$ & 0.9 & 0.8 & 0.8 & 0.8 \\
		\hline
	\end{tabular}
	
\end{table}

\end{document}